\documentclass[runningheads]{llncs}

\usepackage{eccv}

\usepackage{eccvabbrv}

\usepackage{url}            
\usepackage{booktabs}       
\usepackage{amsfonts}       
\usepackage{pifont}
\newcommand{\cmark}{\ding{51}}%
\newcommand{\xmark}{\ding{55}}%
\usepackage{fontawesome}
\usepackage{nicefrac}       
\usepackage{microtype}      
\usepackage{tikz}
\usepackage{xcolor}         
\usepackage{amsmath}
\usepackage{amssymb}
\usepackage{bm}
\usepackage{adjustbox}
\usepackage{dashrule}
\usepackage{mathtools}
\usepackage{makecell}
\usepackage{colortbl}
\usepackage{graphicx}
\usepackage{multirow}

\usepackage{wrapfig}
\usepackage{tikz}
\usepackage{cite}
\usepackage{float}
\usepackage{enumitem}
\usepackage{dsfont}
\usepackage{array}
\newcolumntype{x}[1]{>{\centering\let\newline\\\arraybackslash\hspace{0pt}}p{#1}}
\usepackage{ctable}

\allowdisplaybreaks

\usepackage[accsupp]{axessibility}  

\definecolor{blueblack}{RGB}{0, 108, 173}
\definecolor{taborange}{RGB}{235, 127, 14}
\definecolor{tabgreen}{RGB}{30, 160, 30}
\definecolor{tabpurple}{RGB}{128, 103, 189}
\definecolor{tabred}{RGB}{214, 39, 40}
\definecolor{tabpink}{RGB}{240, 20, 255}
\definecolor{tabblue}{RGB}{14, 22, 255}
\definecolor{blueviolet}{RGB}{138,43,226}
\definecolor{forestgreen}{RGB}{34,139,34}

\usepackage[most]{tcolorbox}
\newtcbox{\boxpurple}[1][]{
  notitle, 
  nophantom,
  nobeforeafter,
  math upper,
  colframe=white, 
  #1,
  on line, 
  boxsep=4pt, left=0pt,right=0pt,top=0pt,bottom=0pt,
  colframe=white,
  colback=blueviolet!30,  
  highlight math style={enhanced}
}

\newtcbox{\boxgreen}[1][]{
  notitle, 
  nophantom,
  nobeforeafter,
  math upper,
  colframe=white, 
  #1,
  on line, 
  boxsep=4pt, left=0pt,right=0pt,top=0pt,bottom=0pt,
  colframe=white,
  colback=forestgreen!30,  
  highlight math style={enhanced}
}

\definecolor{sol_light_blue}{RGB}{38, 139, 210}
\definecolor{sol_blue}{RGB}{38, 139, 210}
\definecolor{nord_blue}{RGB}{38, 139, 210}
\definecolor{sol_green}{RGB}{163, 190, 140}
\definecolor{sol_red}{RGB}{220, 50, 47}
\definecolor{nord_red}{RGB}{250, 190, 192}
\definecolor{nord_green}{RGB}{182, 215, 168}
\definecolor{nord_yellow}{RGB}{255, 229, 153}

\definecolor{beer_orange}{RGB}{242, 142, 28}

\DeclareMathOperator*{\argmin}{arg\,min}

\usepackage[pagebackref,breaklinks,colorlinks,citecolor=eccvblue]{hyperref}

\usepackage{orcidlink}

\begin{document}

\title{Fast Kernel Scene Flow} 


\author{Xueqian Li \and
Simon Lucey}

\authorrunning{X.~Li~\etal}

\institute{Australian Institute for Machine Learning \\
The University of Adelaide \\
\email{xueqian.li@adelaide.edu.au}\\
\url{https://github.com/Lilac-Lee/FastKernelSF}}

\maketitle

\begin{abstract}
In contrast to current state-of-the-art methods, such as NSFP \cite{li2021neural}, which employ deep implicit neural functions for modeling scene flow, we present a novel approach that utilizes classical kernel representations. 
This representation enables our approach to effectively handle dense lidar points while demonstrating exceptional computational efficiency---compared to recent deep approaches---achieved through the solution of a linear system.
As a runtime optimization-based method, our model exhibits impressive generalizability across various out-of-distribution scenarios, achieving competitive performance on large-scale lidar datasets.
We propose a new positional encoding-based kernel that demonstrates state-of-the-art performance in efficient lidar scene flow estimation on large-scale point clouds.
An important highlight of our method is its near real-time performance ($\sim$150-170 ms) with dense lidar data ($\sim$8k-144k points), enabling a variety of practical applications in robotics and autonomous driving scenarios.
\end{abstract}

\section{Introduction}
\label{sc:intro}

In robotics and autonomous driving scenarios, scene flow has emerged as an essential task for various applications, including point cloud densification~\cite{li2021neural, wang2022neural, huang2022dynamic}, providing correspondence for dynamic objects~\cite{yang2023emernerf}, and serving as motion cues to open-world perceptions~\cite{najibi2022motion}.
Consequently, scene flow estimation stands as a cornerstone in understanding complex and dynamic scenes.
Current state-of-the-art scene flow estimation methods can be summarized by three key properties (as shown in~\cref{fig:figure_01}):

\begin{figure}[t]
    \centering
    \includegraphics[width=0.65\linewidth]{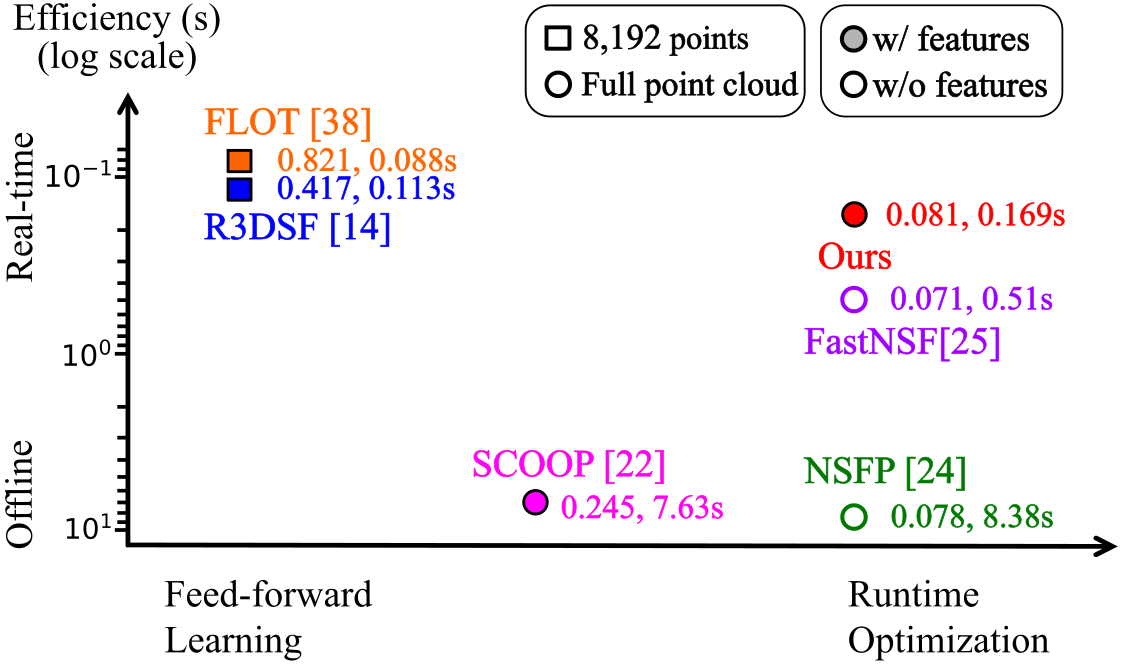}
    \caption{Current point cloud-based scene flow methods can be summarized based on three key properties: 
    (1) whether they are feed-forward learning or runtime optimization (x-axis); (2) whether they leverage point features or not (shaded or empty legends); (3) whether they can be applied in real-time, or are preferable for offline applications (y-axis). 
    Feed-forward learning, such as FLOT~\cite{puy20flot} and R3DSF~\cite{gojcic2021weakly}, learn features from data during training.
    Typically, these methods are applied to sparse, small-scale datasets (8,192 points), and exhibit inferior performance when tested on out-of-distribution data.
    In contrast, the runtime optimization-based method NSFP~\cite{li2021neural} is dominated in dense lidar flow estimation but suffers from extremely slow computation (8.38s).
    FastNSF~\cite{li2023fast} addresses this issue and achieves up to 30 times speedups (0.51s).
    As a hybrid method, SCOOP~\cite{lang2023scoop} learns features for point correspondence but still faces computational inefficiency (7.63s).
    Our method integrates a per-point embedding-based feature within a kernel representation that solves a linear system, achieving near real-time performance (0.169s) with an end-point error of 0.081 on dense lidar points.
    }
    \vspace{-0.3cm}
    \label{fig:figure_01}
\end{figure}

\vspace{0.2cm}
\noindent\textit{\textbf{(1) Feed-forward vs. runtime optimization.}}\,\,
Traditional scene flow estimation formulates the non-rigid registration problem to explicitly estimate flow vectors from raw points~\cite{chui2003new, pauly2005example, amberg2007optimal}.
In contrast, feed-forward learning methods~\cite{liu2019flownet3d, puy20flot, wu2020pointpwc, gojcic2021weakly, kittenplon2020flowstep3d} adopt conventional training pipelines where deep neural networks are optimized during training to directly predict flow during inference. 
While these learned methods exhibit relatively fast speed during inference, they often show inferior performance on out-of-distribution (OOD) datasets~\cite{pontes2020scene, li2021neural, najibi2022motion, dong2022exploiting, jin2022deformation, li2023fast} due to their data-dependent nature.
Moreover, learned methods face challenges in scaling up to handle dense, large-scale datasets.
As a result, benchmarking of learned methods primarily focuses on sparse, small-scale, synthetic datasets such as FlyingThings3D~\cite{mayer2016large} and strictly preprocessed datasets~\cite{li2023fast, chodosh2023re}, such as the KITTI scene flow dataset~\cite{menze2015joint, liu2019flownet3d}.

Recently, the vision community has witnessed a resurgence of interest in runtime optimization, leading to significant progress in novel view synthesis~\cite{mildenhall2020nerf}, image reconstruction~\cite{sitzmann2020implicit}, and shape representation~\cite{park2019deepsdf},~\etc.
Of particular note is Neural Scene Flow Prior (NSFP)~\cite{li2021neural} which employs a deep ReLU-MLP to represent dense, large-scale lidar flow through runtime optimization, and improves the state-of-the-art performance of learned methods by a large margin.

\vspace{0.2cm}
\noindent\textit{\textbf{(2) Feature embeddings vs. raw points.}}\,\,
Data-driven learning methods require training on extensive data with flow supervision to obtain high-dimensional feature embeddings of points. 
By contrast, runtime optimization methods work effectively on raw points without the need for any off-line learning or training.
Noteworthy examples such as NSFP~\cite{li2021neural} and its variants~\cite{li2023fast, vidanapathirana2023multi, wang2022neural, chodosh2023re, pontes2020scene} leverage a deep network structure to regularize the predicted flow without relying on explicit point feature embeddings.
A recent work SCOOP~\cite{lang2023scoop} integrates point features by learning a point correspondence model while optimizing flow at runtime.
However, this hybrid pipeline may overlook the potential benefits of integrating feature embeddings during runtime.

\vspace{0.2cm}
\noindent\textit{\textbf{(3) Real-time vs. offline inference.}}\,\,
Feed-forward learning methods for scene flow are typically fast during inference (in hundreds of milliseconds).
Despite the success of NSFP and its variants~\cite{pontes2020scene, li2021neural, vidanapathirana2023multi, chodosh2023re, wang2022neural} runtime optimization methods suffer a fundamental drawback of low computation speed (sometimes orders of magnitude slower than feed-forward learning methods) limiting its application to offline tasks such as integrating long-term flow~\cite{li2021neural, wang2022neural}, providing flow supervision~\cite{chodosh2023re, vedder2023zeroflow}, and pre-training open-world vision systems~\cite{najibi2022motion}.
These computational overheads emerge from the architecture of the deep implicit network and the exhaustive point correspondence search involved in the Chamfer distance loss~\cite{fan2017point}.
FastNSF~\cite{li2023fast} partially addresses the efficiency issue by introducing a distance transform-based loss, achieving speedups of up to 30 times.

\vspace{0.2cm}
\noindent\textit{\textbf{Contributions.}}\,\,
Building upon the aforementioned insights and motivations, we introduce Per-Point Embedding (PPE) kernel scene flow, a novel kernel learning approach for lidar scene flow estimation.
Our method leverages PPE features and solves a linear system at runtime, allowing for fast convergence ($\sim$150-170 ms on Argoverse and Waymo Open full lidar point datasets)
and excellent scalability for dense, large-scale lidar data.
Distinguishing itself from NSFP-based approaches, our kernel-based method optimizes only a linear coefficient vector, whose size equals the number of supporting points.
This results in a compact and efficient formulation, achieving near real-time performance even on dense lidar points ($\sim$8k-144k points).
Furthermore, our method implicitly embeds point features into the kernel function using PPEs, enabling better expressibility.
In line with conclusions from~\cite{zheng2023robust}, we find that analytical PPE features, such as Random Fourier Feature (RFF)-based positional encoding, serve as robust surrogates for learned embeddings while maintaining great scalability to dense lidar points.

To summarize, our method enjoys the following properties:
\vspace{-0.2cm}
\begin{itemize}[label={--}, leftmargin=1.5em]
    \item Our method employs a classical kernel representation that exclusively optimizes a linear coefficient vector at runtime, maintaining out-of-distribution generalizability akin to the runtime optimization of a deep implicit network, and achieving competitive performance on large-scale lidar datasets.
    \item The kernel representation enables the implicit embedding of point features. 
    We therefore investigate raw points, learned features, and analytical positional encodings to incorporate robust PPEs.
    \item Our linear optimization model demonstrates effective scalability to dense lidar point clouds, achieving near real-time performance.
\end{itemize}

\section{Related Work}
\label{sc:related}
\noindent\textit{\textbf{Scene flow estimation.}}\,\,
Scene flow is a crucial task in autonomous driving scenarios, particularly with dynamic lidar scenes.
Recent advances in both full learning-based~\cite{liu2019flownet3d, liu2019meteornet, gu2019hplflownet, wang2020flownet3d++, puy20flot, kittenplon2020flowstep3d, wang2021festa, wu2020pointpwc} and self-supervised learning-based methods~\cite{gojcic2021weakly, wu2020pointpwc, mittal2020just, tishchenko2020self, baur2021slim, kittenplon2020flowstep3d, shen2023self, li2022rigidflow} have shown promising performance on synthetic FlyingThings3D~\cite{mayer2016large} and KITTI scene flow dataset~\cite{menze2015joint}.
Despite these achievements, their performance tends to fall short with OOD data~\cite{pontes2020scene, li2021neural}, posing challenges for direct application to real-world large-scale tasks.

An emerging trend of work focusing on runtime optimization~\cite{pontes2020scene, li2021neural, li2023fast, vidanapathirana2023multi, chodosh2023re} demonstrates outstanding performance with superior generalizability across various real-world datasets.
Notably, NSFP~\cite{li2021neural} pioneered the use of neural network as an implicit prior for representing scene flow.
The remarkable scalability and the regularization of the deep ReLU-MLP contribute to its exceptional performance on large-scale lidar points.
However, a notable limitation of NSFP and its variants is its extremely slow computation time, hindering its real-time applications.
FastNSF~\cite{li2023fast} analyses the two computation overheads that emerged in NSFP and proposes to use distance transform~\cite{breu1995linear, danielsson1980euclidean, borgefors1996digital, bailey2004efficient}, achieving ${\sim}30$ times speedup.
ZeroFlow~\cite{vedder2023zeroflow} uses NSFP for the teacher model and a feedforward network as the student model to distill a fast model.
Additionally, SCOOP~\cite{lang2023scoop} trains a point feature-based correspondence model as flow initialization and refines the flow estimation at runtime.
Analogous to the runtime optimization pipeline, we propose to use PPE kernel method that combines both the generalizability of non-learning methods and the expressibility of point features.

\vspace{0.2cm}
\noindent\textit{\textbf{Kernel learning.}}\,\,
The kernel method initially gained in popularity when introduced in support vector machines (SVMs)~\cite{Cortes1995support}, leveraging the ``kernel trick'' to learn non-linear relationships between data points through a linear solver.
Over time, the kernel method has evolved beyond SVMs and plays foundational roles in many learning algorithms
for classification, regression, clustering, time series analysis, natural language processing,~\etc.
Among different types of kernel functions the radial basis function (RBF) remains the most common and effective, especially for low-dimensional regression and learning problems~\cite{buhmann2000radial}.
Recently, learned feature-based kernels have been explored in 3D surface reconstruction~\cite{williams2022neural, huang2023neural}, introduced as data-dependent kernels.
In this paper, we explore the use of a novel PPE kernel to represent dense, large-scale scene flow.

\section{Method}
\label{sec:approach}

\subsection{Preliminary}
\label{sec:preliminary}

\noindent\textit{\textbf{Scene flow optimization.}}\,\,
Scene flow describes the motion in the 3D space.
Imagine in the autonomous driving scenario, a lidar sensor is collecting point cloud sweeps while the AV is moving.
At time $t\text{-}1$, we capture a point cloud denoted as source $\mathcal{S}_1$.
Later at time $t$, another point cloud is captured through the sensor as target $\mathcal{S}_2$.
We then model the translation of each point $\mathbf{p} \,{\in}\, \mathcal{S}_1$ as flow $\mathbf{f}\,{\in}\, \mathbb{R}^{3}$.
The deformed source point is denoted as $\mathbf{p}' \,{=}\, \mathbf{p} \,{+}\, \mathbf{f}$.
We collect all translational vectors and get scene flow $\mathcal{F} \,{=}\, \{\mathbf{f}_i\}_{i=1}^{|\mathcal{S}_1|}$.

In the conventional optimization algorithm, a distance between the source and the target point cloud is optimized to get the predicted scene flow as
\begin{align}
    \mathcal{F}^* = \argmin_{\mathcal{F}} \sum_{\mathbf{p} \in \mathcal{S}_1} \mbox{D} \left( \mathbf{p}+\mathbf{f}, \mathcal{S}_2 \right) + \lambda \mbox{C}(\mathbf{f}) \,, 
    \label{eq:optim_02}
\end{align}
where $\mbox{D}$ is a point distance function, such as the Chamfer distance~\cite{fan2017point}; $\mbox{C}$ is a regularizer term, such as a Laplacian regularizer; $\lambda$ is a coefficient of the regularizer.

\vspace{0.2cm}
\noindent\textit{\textbf{Neural scene flow prior.}}\,\,
Inspired by the traditional optimization-based method, neural scene flow prior (NSFP)~\cite{li2021neural} optimizes a neural representation of the flow at runtime.
Instead of directly optimizing the scene flow with an explicit regularizer, NSFP optimizes the parameters $\mathbf{\Gamma}$ of a coordinate network $g$---usually a ReLU-MLP---with an implicit regularizer on itself:
\begin{align}
    \mathbf{\Gamma}^* = \argmin_{\mathbf{\Gamma}} \sum_{\mathbf{p} \in \mathcal{S}_1} \mbox{D} \left( \mathbf{p} + g \left(\mathbf{p}; \mathbf{\Gamma} \right), \mathcal{S}_2 \right) \,. 
    \label{eq:optim_main}
\end{align}
The optimized flow can be represented as $\mathbf{f}^{*} \;{=}\; g \left(\mathbf{p};\, \mathbf{\Gamma}^{*} \right)$.

Similar to traditional optimization, the distance function can be Chamfer loss that deals with different numbers of points in the source and the target ($|\mathcal{S}_1| \,{\neq}\, |\mathcal{S}_2|$) as
\begin{align}
    \mbox{D} \left(\mathbf{p},\mathcal{S} \right) = \min_{\mathbf{x} \in \mathcal{S}} \lVert \mathbf{p} - \mathbf{x} \rVert_2^2 \,. 
    \label{eq:chamfer_losss}
\end{align}

\vspace{0.2cm}
\noindent\textit{\textbf{Kernel method.}}\,\,
The kernel method is widely used to represent the complex and non-linear relationship of data by solving linear functions.
A common kernel representation can be denoted as $y \,{=}\, \bm{\alpha} \mathcal{K} (\mathbf{x}, \mathbf{x^\prime})$, where $\bm{\alpha}$ are linear coefficients.
The key advantage of kernel learning lies in its implicit representation of the data through the kernel function $\mathcal{K}\,{=}\,\langle \mathbf{x}, \mathbf{x}^\prime \rangle$, an inner product of data point.
To allow for a more structured representation, a feature mapping $\phi\,{:}\, \mathcal{X} \,{\rightarrow}\, \mathcal{V}$ is applied to transform the input data to a high-dimensional feature space.
Recent research~\cite{williams2022neural, huang2023neural} further explore data-dependent kernels through the learning of deep neural networks, enabling more expressibility.

\begin{figure}[t]
    \centering
    \includegraphics[width=\linewidth]{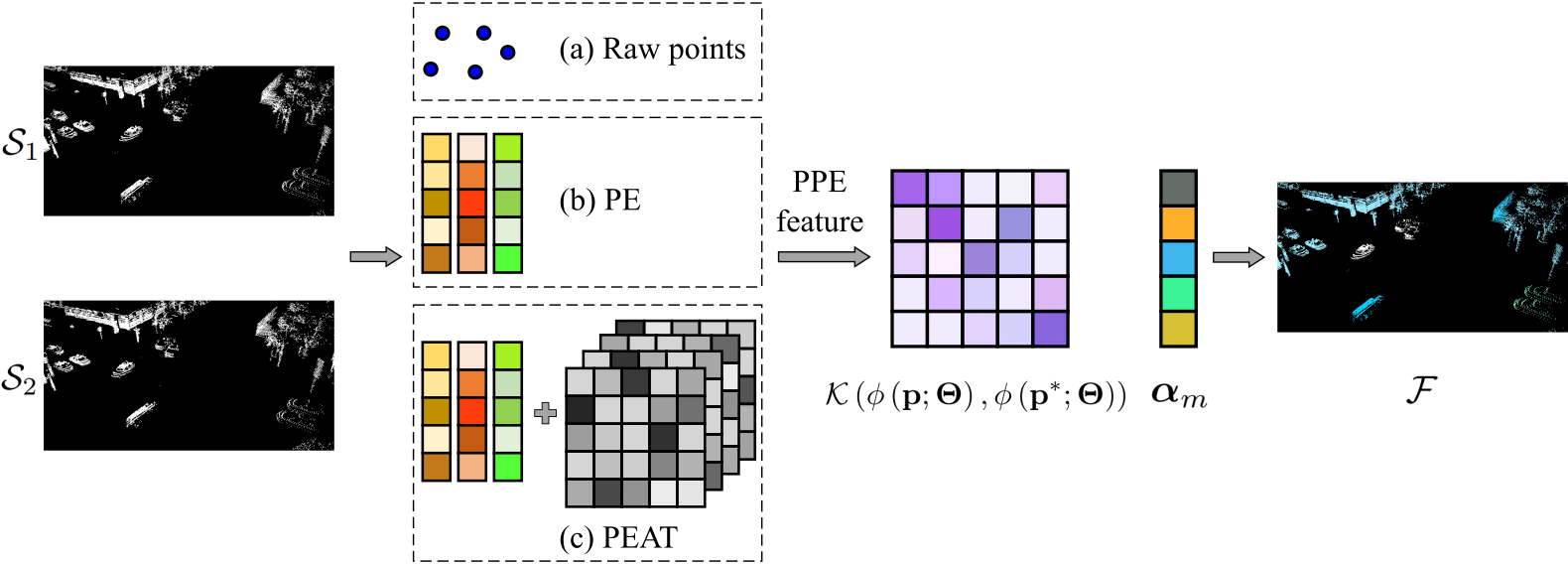}
    \caption{Framework of PPE kernel scene flow.
    With input point cloud $\mathcal{S}_1$, $\mathcal{S}_2$, we could use different approaches to extract per-point features.
    For example:
    (a) Raw points, which treat original points as point features; 
    (b) PE, which uses an RFF-based positional encoding to embed the input point to high-frequency features; 
    (c) PEAT, which extracts learned point features using positional encoding and self-attention.
    A kernel function $\mathcal{K}$ is then employed to compute the similarity matrix between these two inputs based on their point features.
    Finally, a linear coefficient vector $\bm{\alpha}$ is optimized per sample to predict the final flow.
    Our model is compact and fast, with only $\bm{\alpha}$ being the learnable parameter.
    The PPE features can either be pre-trained offline or analytical positional encodings.
    }
    \vspace{-0.3cm}
    \label{fig:kernel}
\end{figure}

\subsection{Kernel Representation for Scene Flow}
\label{sec:kernel_sf}

Our model predicts the scene flow from the raw input of 3D point coordinates.
Unlike NSFP~\cite{li2021neural} and its variants~\cite{li2023fast, vidanapathirana2023multi} which use a deep neural network as implicit prior, we explore the kernel representation of scene flow as
\begin{align}
     \mathbf{f} & = \sum_{m}^{M} \bm{\alpha}_m \mathcal{K} \left(\mathbf{p}, \mathbf{p}^* \right) \,,
     \label{eq:flow_kernel}
\end{align}
where $\bm{\alpha}_m$ denotes linear coefficients, and $M$ is the number of supporting points $\mathbf{p}^*$.
$\mathcal{K} \left(\mathbf{p}, \mathbf{p}^* \right)$ is the kernel basis function on raw points $\mathbf{p}$. 
Here we use the common radial basis function (RBF) kernel: 
\begin{align}
    \mathcal{K}(\mathbf{p}, \mathbf{p}^*) = \exp{\left( -\frac{\lVert \mathbf{p} - \mathbf{p}^* \rVert^2}{2 \sigma^2} \right)} \,. 
    \label{eq:rbf_kernel}
\end{align}
Other basis functions, such as Sinc, Laplacian,~\etc, will be discussed in~\cref{sec:exp:different_kernel}.

Instead of directly optimizing for explicit flow as in traditional optimizations, we optimize kernel coefficients $\bm{\alpha}$ through the point distance function as
\begin{align}
    \bm{\alpha}^* {=} \argmin_{\bm{\alpha}} \sum_{\mathbf{p} \in \mathcal{S}_1} \mbox{D} \left( \mathbf{p} {+} \sum_m^M \boxpurple{\bm{\alpha}_m} \boxgreen{\mathcal{K}(\mathbf{p}, \mathbf{p}^*)}, \mathcal{S}_2 \right) {+} \lambda \left| \boxpurple{\bm{\alpha}} \right| \, , 
    \label{eq:optimize_kernel}
\end{align}
where we add L1-regularizer to the coefficients.
Here, only the purple-shaded parts are optimized during runtime, the green-shaded parts are fixed during optimization.
An illustration of our method is shown in~\cref{fig:kernel}.

\setlength{\columnsep}{10pt}%
\begin{wrapfigure}[15]{r}{0.45\textwidth}
    \vspace{-0.55cm}
    \centering
    \includegraphics[width=0.55\linewidth]{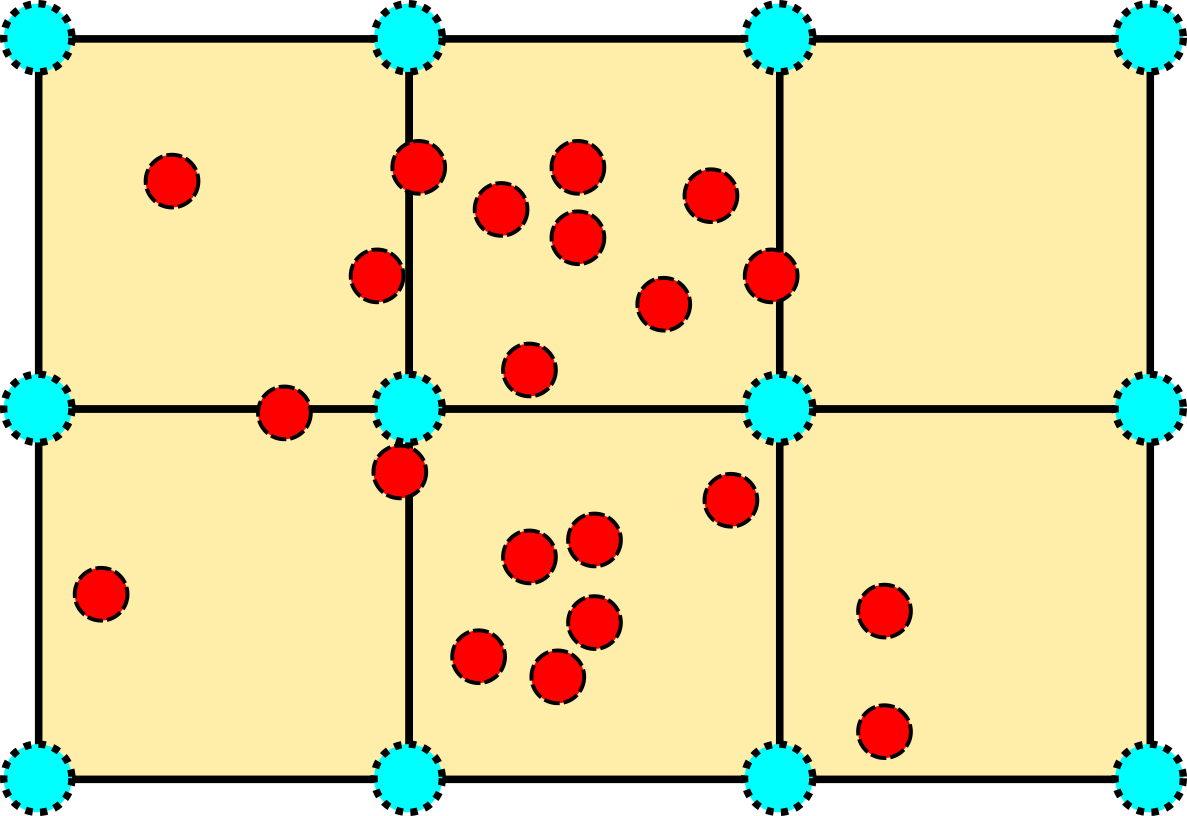}
    \caption{The \textbf{\color{red}red} points indicate raw point cloud while the \textbf{\color{cyan}cyan} points represent supporting grid points.
    $\mathbf{p}$ and $\mathbf{p}^*$ do not indicate necessary correspondence between them.
    For example,~\cref{eq:rbf_kernel} describes a kernel function of size $N \,{\times}\, M$, where $N$ is the number of points in the source point cloud, and $M$ is the number of grid points.}
    \label{fig:grid_point}
\end{wrapfigure}
\vspace{0.2cm}
\noindent\textit{\textbf{Grid points as support.}}\,\,
In~\cref{eq:flow_kernel,eq:rbf_kernel,eq:optimize_kernel}, the supporting point $\mathbf{p}^*$ could be selected from target point cloud $\mathcal{S}_2$.
However, given the fact that lidar points are noisy and have various densities, dense in regions with a large number of points and sparse in areas with many empty spaces, we use regular grid points $\mathbf{p}^*$ to replace the target point.
We show a 2D illustration in~\cref{fig:grid_point}.
The size of the regular grid points $M$ is smaller than the number of target points.
This grid point formulation is robust against noisy data points and is computationally efficient,
especially when dealing with dense lidar points.

\vspace{0.2cm}
\noindent\textit{\textbf{Difference to neural priors.}}\,\,
Although utilizing runtime optimization, our method is fundamentally different from neural prior-based methods~\cite{li2021neural, li2023fast, vidanapathirana2023multi}.
Instead of optimizing a deep neural network,~\ie, 8-layer ReLU-MLPs, our method only optimizes a linear coefficient vector, which is equivalent to a single-layer linear network without any non-linear activations.

\subsection{Per-Point Embedding Kernel}
\label{sec:ppe_kernel}
To obtain a more expressible data-dependent kernel, we add a per-point embedding (PPE)-based feature to~\cref{eq:flow_kernel} as
\begin{align}
    \mathbf{f} & = \sum_m^M \bm{\alpha}_m \mathcal{K} \left( \phi\left( \mathbf{p}; \mathbf{\Theta} \right), \phi \left( \mathbf{p}^*; \mathbf{\Theta} \right) \right) \,,
    \label{eq:feature_kernel}
\end{align}
where $\phi(\cdot)$ denotes per-point embeddings.
We use $\mathcal{K}_{\mathbf{\Theta}}$ for brevity to denote this feature-based kernel.
In practice, we explore different PPEs and focus on raw points, learned point features, and analytical positional encodings (PEs).

\vspace{-0.4cm}
\subsubsection{3.3.1~~Raw Points}\mbox{}\label{sec:method:raw_points}
\vspace{0.1cm}

\noindent Since each point in a point cloud represents the positional/spatial information, we directly use raw points to compute the kernel as
\begin{align}
    \mathcal{K}_{\mathbf{\Theta}} & = \mathcal{K}_{\mathbb{I}} = \mathcal{K} \left( \phi\left( \mathbf{p}; \mathbb{I} \right), \phi \left( \mathbf{p}^*; \mathbb{I} \right) \right) = \mathcal{K} \left( \mathbf{p}, \mathbf{p}^* \right) \,,
    \label{eq:raw_point}
\end{align}
where $\mathbb{I}$ denotes the identity matrix.

\vspace{-0.4cm}
\subsubsection{3.3.2~~Learned PEAT Features}\mbox{}\label{sec:method:peat}
\vspace{0.1cm}

\noindent Transformers have proven effective in extracting point cloud features~\cite{zhao2021point, guo2021pct}.
PEAT~\cite{zheng2023robust} proposes to use positional embedding to replace the per-point embedding in the point cloud transformer, resulting in more robust point features.
We explore the use of PEAT as a per-point embedding as
\begin{align}
    \phi(\mathbf{p}; \mathbf{\Theta}) = & \mbox{PEAT} (\mathbf{p}; \mathbf{\Theta}) = \mbox{Softmax} \left( \mathbf{X} \mathbf{W}_Q \mathbf{W}_K^T \mathbf{X}^T \right) \mathbf{X} \mathbf{W}_V \,, \label{eq:peat}
\end{align}
where $\mathbf{X} \,{=}\, \mbox{PE} (\mathbf{p})$, $\mathbf{W}_Q$, $\mathbf{W}_K$, $\mathbf{W}_V$ are transformer weights parameterized by $\mathbf{\Theta}$.

\vspace{0.2cm}
\noindent\textit{\textbf{K-NN-based PEAT features.}}\,\,
Scalability remains a challenge when applying attention-based features to dense point clouds.
Approaches such as~\cite{hui2021pyramid, zhang2022patchformer, park2022fast, qin2022geometric} address this issue through patch-based or graph-based local feature embeddings.
On the other hand,~\cite{zhao2021point, guo2021pct} utilize k-nearest neighbors (k-NN) to subsample point features.
However, these feature aggregation strategies alter the number of points, making them unsuitable for per-point flow estimation.

Motivated by these previous works, we propose to use k-NN to aggregate attention maps within local neighboring points.
Specifically, we sample $L$ number of points using k-NN for each input point as
\begin{align}
    \mathbf{p}_L = \eta \left( \mathbf{p}; \mathcal{S}; L \right)  \,,\label{eq:knn_sample}
\end{align}
where $\eta(\cdot)$ is the sampling function, $\mathbf{p} \,{\in}\, \mathcal{S}$.
Therefore, the positional embedding of $\mathbf{p}_L$ becomes $\mathbf{X}_L \,{=}\, \mbox{PE}(\mathbf{p}_L)$.
The PEAT-KNN feature is then defined as
\begin{align}
     \mbox{PEAT-KNN}(\mathbf{p}; \mathbf{\Theta}) = \mbox{Softmax} \left( \mathbf{x} \mathbf{W}_Q \mathbf{W}_K^T \mathbf{X}_L^T \right) \mathbf{X}_L \mathbf{W}_V \,,~~\forall \mathbf{x} \in \mathbf{X}. 
    \label{eq:knn_peat}
\end{align}

This formulation significantly reduces the size of attention maps, allowing for a more compact representation using PEAT features for dense lidar points.

\vspace{0.2cm}
\noindent\textit{(1) Supervised learning.}
\vspace{0.1cm}

\noindent Given that synthetic datasets provide ground truth flow, we can train a supervised model end-to-end with optimizing network parameters $\mathbf{\Theta}$ along with the coefficients $\bm{\alpha}$ as
\begin{align}
    \argmin_{\bm{\alpha}, \mathbf{\Theta}} \left\Vert \mathbf{f}_{gt} - \sum_m^M \bm{\alpha}_m \mathcal{K}_{\mathbf{\Theta}} \right\Vert^2_2 + \lambda \left| \bm{\alpha} \right| \,.
    \label{eq:supervised_learning_01}
\end{align}

\vspace{0.2cm}
\noindent\textit{(2) Self-supervised learning.}
\vspace{0.1cm}

\noindent When the training labels are not available, we could easily adapt the objective to self-supervised training as
\begin{align}
    \argmin_{\bm{\alpha}, \mathbf{\Theta}} \sum_{\mathbf{p} \in \mathcal{S}_1} \mbox{D} \left( \mathbf{p} {+} \sum_m^M \bm{\alpha}_m \mathcal{K}_{\mathbf{\Theta}}, \mathcal{S}_2 \right) {+} \lambda \left| \bm{\alpha} \right| . 
    \label{eq:self_supervised_learning}
\end{align}

The network parameter $\mathbf{\Theta}$ will be optimized across all data samples, while the kernel coefficient $\bm{\alpha}$ will be optimized per sample.

\vspace{-0.4cm}
\subsubsection{3.3.3~~Positional Encoding}\mbox{}\label{sec:method:pe}
\vspace{0.1cm}

\noindent According to~\cite{zheng2023robust}, positional encoding (PE) is a robust surrogate to learned per-point embeddings, in a way, they both transform the low-dimensional input data to a high-dimensional feature space.
However, the intrinsic difference is that PE is an analytical embedding, and only one hyperparameter---the bandwidth---needs to be tuned~\cite{zheng2022trading}.

Another notable advantage of PE is its scalability which allows the representation of large-scale data in a compact form.
Therefore, we explore positional encoding as a feature embedding to integrate with kernel methods as
\begin{align}
    \mathcal{K}_{\mathbf{\Theta}} & = \mathcal{K}_{\beta} = \mathcal{K} \left( \phi\left( \mathbf{p}; \beta \right), \phi \left( \mathbf{p}^*; \beta \right) \right) \,.
    \label{eq:pe_kernel}
\end{align}
Here, we use the random Fourier feature (RFF)~\cite{tancik2020fourier} with a scale parameter $\beta$.

\section{Experiments}
\label{sec:exp}

We compared scene flow results across two large-scale autonomous driving datasets against state-of-the-art non-learning methods NSFP~\cite{li2021neural}, FastNSF~\cite{li2023fast}, hybrid learning method SCOOP~\cite{lang2023scoop}, self-supervised learning method R3DSF~\cite{gojcic2021weakly}, and full-supervised learning method FLOT~\cite{puy20flot}.

\vspace{0.2cm}
\noindent\textit{\textbf{Datasets.}}\,\,
Considering that scene flow is one of the crucial tasks for autonomous vehicles, we focus on two prominent autonomous driving datasets Argoverse~\cite{chang2019argoverse} and Waymo Open~\cite{sun2020scalability}.
These two datasets contain large-scale, dynamic, real-world, challenging lidar scenes, and have drawn wide attention recently in the scene flow community~\cite{pontes2020scene, li2021neural, li2023fast, vidanapathirana2023multi, wang2022neural, gojcic2021weakly}.
We use the pre-processed Argoverse and Waymo Open scene flow data released by~\cite{li2023fast} for testing, and the FlyingThings3D~\cite{mayer2016large} dataset used in~\cite{liu2019flownet3d, gojcic2021weakly, li2021neural, puy20flot, lang2023scoop} for training.

\vspace{0.2cm}
\noindent\textit{\textbf{Metrics.}}\,\,
We set up performance metrics as defined in~\cite{liu2019flownet3d}, widely used across various scene flow methods.
\textbf{(1) 3D End-Point Error (EPE) $\mathcal{E}(m)$} quantifies the mean absolute distance between two point clouds; 
\textbf{(2) Strict accuracy $Acc_5(\%)$} is defined by the absolute EPE $\mathcal{E} < 0.05$m or the relative EPE $\mathcal{E}^{\prime} < 5\%$; 
\textbf{(3) Relaxed accuracy $Acc_{10}(\%)$} denotes absolute EPE $\mathcal{E} < 0.1$m or the relative EPE $\mathcal{E}^{\prime} < 10\%$; 
\textbf{(4) Angle error $\theta_\epsilon(rad)$} evaluates the mean angle error between the prediction and the ground truth flow.
We also add the \textbf{(5) Computation time $T(s)$} of total optimization
as an efficiency measurement.

\vspace{0.3cm}
\noindent\textit{\textbf{Implementation details.}}\mbox{}

\vspace{0.1cm}
\noindent\textbf{(1) Ours:}
We implemented our PPE kernel methods using PyTorch~\cite{paszke2019pytorch} with Adam~\cite{kingma2014adam} optimizer.
Our method has three variants: \textbf{Ours (Point)} with raw points, \textbf{Ours (PEAT/PEAT-KNN)} with learned attention-based features, and \textbf{Ours (RFF)} with RFF-based positional embeddings.
For a fair comparison with NSFP~\cite{li2021neural} and FastNSF~\cite{li2023fast}, we used both Chamfer and DT loss.

We train the PEAT/PEAT-KNN kernel using the synthetic FlyingThings3D dataset.
As discussed in~\cref{sec:method:peat}, the learned PEAT feature is not scalable to dense points.
Therefore, when evaluated on the full point datasets, we used the PEAT-KNN strategy.
However, for evaluations on 8,192 points, we keep the original PEAT feature.
It is worth noting that for PEAT-based kernels, the original target points are used as supporting points, given that the point feature was trained on raw point clouds.
For other types of kernels,~\ie, point kernel and RFF kernel, we use grid points as supporting points.

\vspace{0.2cm}
\noindent\textbf{(2) Baselines:}
For performance comparison with other methods---\textbf{NSFP}~\cite{li2021neural}, \textbf{FastNSF}~\cite{li2023fast}, \textbf{SCOOP}~\cite{lang2023scoop}, \textbf{FLOT}~\cite{puy20flot}, and \textbf{R3DSF}~\cite{gojcic2021weakly}, we adopted official implementations released by the authors.
Specifically, for NSFP~\cite{li2021neural}, we adopted the baseline version implemented in FastNSF~\cite{li2023fast} for a faster convergence ($\sim$2 times speedups).
All the learning-based methods and the hybrid method were trained on the FlyintThings3D dataset, and directly tested on the lidar data.

We run all experiments on a single RTX 3090Ti GPU with CUDA 11.6.

\begin{table*}[t]
\caption[]{\textbf{Performance on Argoverse scene flow dataset.}
We tested with 212 scene flow examples from the Argoverse validation dataset, presenting the results in two sections. 
The upper table shows results for the full point cloud, with the number of points ranging from 30k to 80k.
In this section, we focus on discussing runtime optimization-based and hybrid methods.
The lower table shows results for sparse point clouds with 8,192 points, where we specifically include results for learning-based methods.
In both tables, performance levels are color-coded as: \colorbox{nord_green}{green box} denotes the best level of performance, \colorbox{nord_yellow}{yellow box} represents the second-best level of performance, and \colorbox{nord_red}{red box} indicates the worst level of performance.
We categorize each method as either a runtime optimization \cmark, a hybrid \faCircleO, or a feed-forward learning method \xmark.
Additionally, we indicate whether the method leverages point features as \cmark or \xmark.
}
    \centering
    \begin{adjustbox}{width=\linewidth}
    \begin{tabular}{@{}clccx{0.13\linewidth}x{0.13\linewidth}x{0.13\linewidth}x{0.13\linewidth}x{0.13\linewidth}c@{}}
        \toprule
        & \thead{\normalsize Method} & \thead{\normalsize Runtime Opt.} & \thead{\normalsize Point Features} &\thead{${\mathcal{E}}{(m)}\downarrow$} &\thead{${Acc_5}{(\%)}\uparrow$} &\thead{${Acc_{10}}{(\%)}\uparrow$} &\thead{${\theta_{\epsilon}}{(rad)}\downarrow$} &\thead{$T(s)\downarrow$} &\\
        \midrule
        & NSFP~\cite{li2021neural} (baseline) & \cmark & \xmark & \cellcolor{nord_green}0.078 & \cellcolor{nord_yellow}\underline{69.46} & \cellcolor{nord_yellow}\underline{86.22} & \cellcolor{nord_green}\textbf{0.253} & \cellcolor{nord_red}8.38 \\
        & SCOOP~\cite{lang2023scoop} & \faCircleO & \cmark & \cellcolor{nord_red}0.245 & \cellcolor{nord_red}45.29 & \cellcolor{nord_red}61.00 & \cellcolor{nord_red}0.422 & \cellcolor{nord_red}7.63 \\
        & FastNSF~\cite{li2023fast} & \cmark & \xmark & \cellcolor{nord_green}\textbf{0.071}& \cellcolor{nord_green}\textbf{80.05} & \cellcolor{nord_green}\textbf{90.71} & 0.289 
        & \cellcolor{nord_yellow}0.51 \\
        & FastNSF~\cite{li2023fast} (linear) & \cmark & \xmark & 0.106 & 65.00 & 82.85 & 0.319
        & \cellcolor{nord_yellow}0.43 \\
        & Ours (Point Kernel, CD) & \cmark & \xmark & \cellcolor{nord_yellow}0.084 & 61.96 & 81.66 & \cellcolor{nord_yellow}0.267 & 3.59 \\
        & Ours (Point, DT) & \cmark & \xmark & \cellcolor{nord_yellow}0.082 & \cellcolor{nord_yellow}68.16 & \cellcolor{nord_yellow}84.89 & 0.298 & \cellcolor{nord_green}\textbf{0.150} \\
        & Ours (PEAT-KNN, CD) & \faCircleO & \cmark & 0.111 & 54.96 & 77.45 & 0.296 & \cellcolor{nord_red}7.23 \\
        & Ours (RFF, CD) & \cmark & \cmark & \cellcolor{nord_green}\underline{0.077} & 64.28 & 83.17 & \cellcolor{nord_green}\underline{0.258} & 3.21 \\
        & Ours (RFF, DT) & \cmark & \cmark & \cellcolor{nord_yellow}0.081 & \cellcolor{nord_yellow}69.45 & \cellcolor{nord_yellow}84.95 & 0.293 & \cellcolor{nord_green}\underline{0.169} \\
        \midrule
        & FLOT~\cite{puy20flot} & \xmark & \cmark & \cellcolor{nord_red}0.821 & \cellcolor{nord_red}2.00 & \cellcolor{nord_red}8.84 & \cellcolor{nord_red}0.967 & \cellcolor{nord_green}0.088 \\
        & R3DSF~\cite{gojcic2021weakly} & \xmark & \cmark & \cellcolor{nord_red}0.417 & \cellcolor{nord_red}32.52 & \cellcolor{nord_red}42.52 & \cellcolor{nord_red}0.551 & \cellcolor{nord_green}\textbf{0.113} \\
        & NSFP~\cite{li2021neural} & \cmark & \xmark & \cellcolor{nord_yellow}0.113 & 46.32 & 72.68 & \cellcolor{nord_green}0.347 & \cellcolor{nord_red}2.86 \\
        & SCOOP~\cite{lang2023scoop} & \faCircleO & \cmark & \cellcolor{nord_red}0.306 & \cellcolor{nord_red}32.78 & \cellcolor{nord_red}54.63 & \cellcolor{nord_red}0.488 & 0.282 \\
        & FastNSF~\cite{li2023fast} & \cmark & \xmark & 0.118 & \cellcolor{nord_green}\textbf{69.93} & \cellcolor{nord_green}\textbf{83.55} & \cellcolor{nord_yellow}0.352 & \cellcolor{nord_green}\underline{0.124} \\
        & Ours (Point, CD) & \cmark & \xmark & 0.128 & 42.32 & 66.86 & \cellcolor{nord_yellow}0.359 & 0.862 \\
        & Ours (Point, DT) & \cmark & \xmark & 0.137 & 45.17 & 68.95 & 0.372 & \cellcolor{nord_yellow}0.147 \\
        & Ours (PEAT, CD) & \faCircleO & \cmark & \cellcolor{nord_yellow}\underline{0.109} & 45.04 & 72.18 & \cellcolor{nord_green}\textbf{0.342} & 2.15 \\
        & Ours (PEAT, DT) & \faCircleO & \cmark & 0.132 & 43.05 & 69.25 & \cellcolor{nord_green}0.348 & 0.270 \\
        & Ours (RFF, CD) & \cmark & \cmark & \cellcolor{nord_green}\textbf{0.099} & \cellcolor{nord_yellow}50.60 & \cellcolor{nord_yellow}74.47 & \cellcolor{nord_green}\underline{0.344} & 0.578 \\
        & Ours (RFF, DT) & \cmark & \cmark & \cellcolor{nord_yellow}0.114 & \cellcolor{nord_yellow}\underline{56.26} & \cellcolor{nord_yellow}\underline{78.07} & 0.375 & \cellcolor{nord_yellow}0.190 \\
        \bottomrule
    \end{tabular}
    \label{tab:main_argoverse}
    \end{adjustbox}
    \vspace{-0.45cm}
\end{table*}

\subsection{Performance Comparison}\label{sec:exp:benchmark}
\begin{figure}[t]
\centering
    \includegraphics[width=\linewidth]{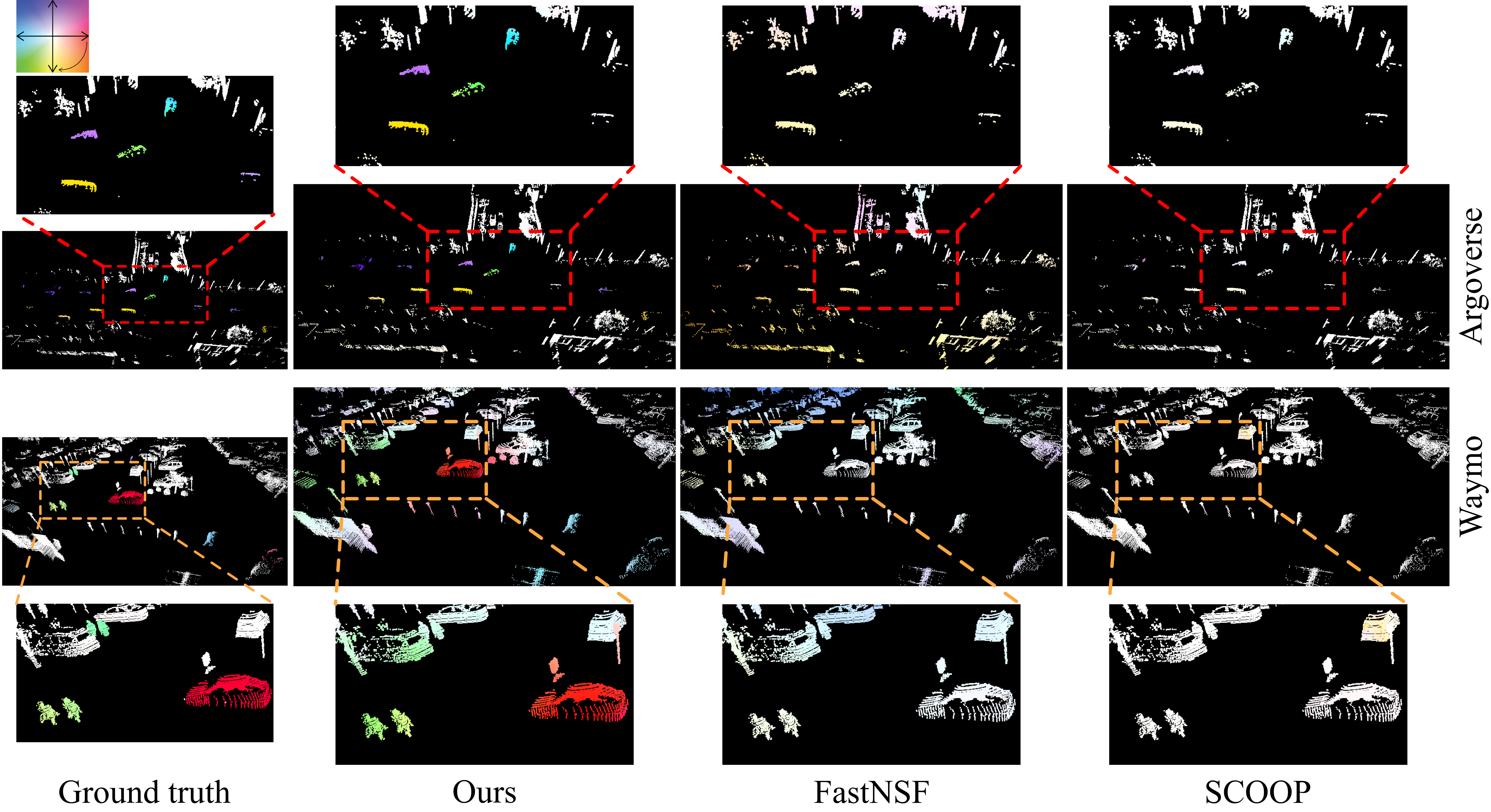}
    \caption{Visual results demonstrate the effectiveness of our method in comparison to FastNSF~\cite{li2023fast} and SCOOP~\cite{lang2023scoop} on two examples from Argoverse and Waymo Open scene flow dataset. 
    The 3D scene flow is presented in a projected 2D view for clearer illustration.
    Zoom-in details are shown for boxed areas.
    In the upper left corner, a color wheel is used to indicate the projected flow magnitude (color intensity) and flow direction (angle). 
    Our method shows great visual results on complicated dynamic AV scenes, successfully capturing both the rigid pose and the dynamic motions.
    FastNSF struggles to capture multiple dynamic objects in some cases, leading to occasional noisy results.
    SCOOP, on the other hand, cannot scale up well to dense points, resulting in near-rigid estimation for most of the scenes.
    } 
    \vspace{-0.3cm}
    \label{fig:argo_qualitative}
\end{figure}

Our main results on the Argoverse scene flow dataset are shown in~\cref{tab:main_argoverse}.
In this table, we show two different density levels of Argoverse lidar data.
The first part of the table shows results tested on the full, dense point clouds, and the second part shows results tested on the 8,192 points to accommodate a fair comparison with learning methods.
The visualization of our method compared to SCOOP and FastNSF is shown in~\cref{fig:argo_qualitative}.

\vspace{-0.4cm}
\subsubsection{4.1.1~~Compare to Runtime Optimization and Hybrid Methods}\mbox{}
\vspace{0.1cm}

\noindent\textit{\textbf{Error and accuracy.}}\,\,
When evaluated on full lidar points, runtime optimization-based methods remain the optimal choice in terms of error and accuracy.
Notably, NSFP, FastNSF, and our RFF-kernel methods demonstrate low end-point error and maintain high accuracy, outperforming the hybrid method SCOOP and our PEAT-KNN kernel.

While FastNSF exhibits high accuracy, it does not confer a significant advantage in terms of end-point error or angle error. 
This can be attributed to the accuracy calculation relying on a threshold that does not consider very high fidelity. 
It turns out that our method exhibits similar or even better performance to FastNSF, as illustrated in~\cref{fig:argo_qualitative}. 
In certain scenarios, FastNSF tends to predict smoother flow, resembling rigid flow, owing to the strong implicit regularization emerging from the network structure and the smooth distance transform map. 
In contrast, our method is more sensitive to noisy points and thus needs robust and explicit regularization.

The hybrid method SCOOP shows relatively poor generalizability, even when flow refinement is optimized during runtime.
Visual results in~\cref{fig:argo_qualitative} further illustrate that the prediction from SCOOP tends to be noisy everywhere.
When incorporating analytical RFF PE-based features, our method shows improved performance, validating the advantage of leveraging analytical point features.

\vspace{0.2cm}
\noindent\textit{\textbf{Computation time.}}\,\,
NSFP is slow in computation due to the runtime optimization of the deep network and the correspondence-based Chamfer loss.
This computational inefficiency becomes particularly prominent with dense points.
The re-implemented baseline NSFP, incorporating a k-d tree-based Chamfer distance, still requires around 8 seconds of computation for a single data, making it unsuitable for real-time applications.

As a hybrid method, SCOOP avoids runtime optimization of the deep network.
However, it encounters computational challenges when extracting point features from dense lidar points.
Moreover, the iterative optimization required for flow refinement adds to the computational load.

FastNSF addresses the computational overhead of Chamfer loss in NSFP by introducing a correspondence-free distance transform loss, achieving a significant efficiency improvement (up to $\sim$30 times speedups for up to $\sim$144k points). 
In contrast, our method is compact and achieves near real-time performance (approximately 150-170ms) on dense lidar points with a distance transform loss, resulting in an overall speedup of NSFP by approximately 55 times.

\vspace{-0.4cm}
\subsubsection{4.1.2~~Compare to Learning-Based Methods}\mbox{}
\vspace{0.1cm}

\noindent\textit{\textbf{Error and accuracy.}}\,\,
Similar to experiments on full point clouds, we observe a consistent trend where runtime optimization-based methods maintain lower error and higher accuracy compared to learning-based methods when tested on 8,192 points.
For learning baselines, we choose FLOT and R3DSF, as many learning methods exhibit unreasonable results when tested out-of-distribution (OOD) with different data distributions, different point densities, or different pre-processing techniques~\cite{li2023fast, pontes2020scene, li2021neural, najibi2022motion, dong2022exploiting, jin2022deformation}.
On the contrary, methods that do not train end-to-end demonstrate excellent generalizability to OOD scenarios, especially for runtime optimization-based approaches.
For example, the hybrid method SCOOP shows acceptable generalizability but inferior performance compared to runtime optimization methods.
When point features are added, models trained end-to-end on FlyingThings3D struggle to generalize these features to OOD lidar scenes. 
However, when employing a hybrid approach (SCOOP and our PEAT kernel) that combines feature learning with runtime optimization, the results remain reasonable, indicating improved accuracy.

\vspace{0.2cm}
\noindent\textit{\textbf{Computation time.}}\,\,
While learning-based methods typically exhibit fast computation times (within hundreds of milliseconds), they are often constrained to small-scale data with specific data characteristics. 
Notably, our method achieves comparable efficiency to learning methods even when tested on dense lidar points.
This observation indicates the strong potential for applying robust PPE kernel to scene flow estimation, particularly in real-time autonomous driving scenarios.

\vspace{0.2cm}
\noindent\textit{\textbf{Performance discrepancy to full point clouds.}}\,\,
The central contribution of the paper is for the dense lidar points, a scenario much more likely to be encountered in the real world. 
Our method exhibits competitive flow error to state-of-the-art methods whilst enjoying a state-of-the-art speedup.
For a fair comparison to learning methods, we also provide results on the less-practical 8k point configuration where the advantage of our method is less stark. 
There are two main reasons.
First, dense points inherently provide stronger regularization when using Chamfer loss or DT loss, consequently leading to better performance~\cite{li2021neural}.
Second, as discussed in FastNSF, the computational overhead in NSFP stems from both Chamfer loss and the deep network structure.
With a large number of points, the disadvantage of optimizing a deep ReLU-MLP network becomes prominent due to the per-point operation.
However, as the point cloud becomes sparser, the disadvantage of using per-point MLP diminishes~\cite{li2023fast}.
By contrast, our method demonstrates great computation efficiency when dealing with dense lidar points, achieved by optimizing a single-layer linear network,~\ie, a kernel coefficient vector.

\begin{table}[t]
    \caption{RFF-based PPE features with different kernel functions.}
    \centering
    \begin{adjustbox}{width=0.73\linewidth}
    \begin{tabular}
    {@{}llx{0.13\linewidth}x{0.13\linewidth}x{0.13\linewidth}x{0.13\linewidth}x{0.13\linewidth}c@{}}
    \toprule
    & \thead{\normalsize Kernel} & \thead{$\mathcal{E}(m)\downarrow$} &\thead{$Acc_5(\%)\uparrow$} &\thead{$Acc_{10}(\%)\uparrow$} & \thead{$\theta_{\epsilon}(rad)\downarrow$} &\thead{$T(s)\downarrow$}
    \\ \midrule
    & RBF & 0.081 & 68.45 & 84.95 & 0.293 & 0.169 &\\
    & Sinc & 0.086 & 64.92 & 84.24 & 0.331 & 0.185 &\\
    & Softmax & 0.108 & 56.83 & 75.41 & 0.260 & 3.16 &\\
    & Sigmoid & 0.087 & 56.11 & 81.64 & 0.327 & 0.411 &\\
    & Tanh & 0.091 & 57.37 & 79.60 & 0.322 & 0.406 &\\
    & Laplacian & 0.077 & 70.77 & 85.98 & 0.294 & 0.270 &\\
    \bottomrule
    \end{tabular}
    \end{adjustbox}
    \label{tb:diff_kernel}
    \vspace{-0.3cm}
\end{table}

\vspace{-0.4cm}
\subsubsection{4.1.3~~Other Comparisons}\mbox{}
\vspace{0.1cm}

\noindent
Additional performance comparisons on the Waymo Open scene flow dataset are presented in the supplementary material due to page limits.
In summary, similar patterns are observed when models are evaluated on the Waymo Open dataset.

\vspace{-0.4cm}
\subsection{Different Kernel Functions}
\label{sec:exp:different_kernel}
\vspace{-0.2cm}
In this work, we use a straightforward RBF-based kernel. 
Nonetheless, the versatility of our approach allows for the substitution of various kernel functions. 
We tested different types of kernel functions using RFF-based point features, and these results are shown in~\cref{tb:diff_kernel}.
In summary, these kernels work similarly well on the dense lidar scene flow task.
However, Softmax kernel is relatively slow due to the large similarity matrix.

\vspace{0.2cm}
\noindent(1) \textbf{Sinc kernel} effectively captures distinct features through its rapid frequency decay, resulting in a fast convergence.
It takes the distance between two features as the input.
The common formulation for the Sinc kernel on two features is given by
\begin{align}
    \mathcal{K}_{\mathbf{\Theta}} = \frac{\mbox{sin} (\pi \mathbf{d})}{\pi \mathbf{d}}, \;\; \mathbf{d} = \sigma \lVert \phi(\mathbf{p}; \mathbf{\Theta}) - \phi(\mathbf{p}^*;\mathbf{\Theta}) \rVert^2 \,,
    \label{eq:sinc_kernel}
\end{align}
where $\sigma$ is the scaling factor for the point distance.

\vspace{0.2cm}
\noindent(2) \textbf{Softmax kernel} is a kernel function that directly applies to the inner product $\langle \cdot \rangle$ of two feature vectors.
With this formulation, the number of hyperparameters is reduced from two (scale of the PE, and scale of the kernel) to one (scale of the PE).
The softmax function serves as a selection matrix that effectively gets the most dominant features from the similarity matrix:
\begin{align}
    \mathcal{K}_{\mathbf{\Theta}} =
    \mbox{Softmax} \left( \vphantom{G^T_T} \langle \phi(\mathbf{p}; \mathbf{\Theta}), \phi(\mathbf{p}^*; \mathbf{\Theta}) \rangle \right) \,.
    \label{eq:softmax_kernel}
\end{align}

\vspace{0.2cm}
\noindent(3) \textbf{Sigmoid kernel} is defined as
\begin{align}
    \mathcal{K}_{\mathbf{\Theta}} =
    \mbox{Sigmoid} \left( \vphantom{G^T_T} \langle \phi(\mathbf{p}; \mathbf{\Theta}), \phi(\mathbf{p}^*; \mathbf{\Theta}) \rangle \right) \,.
\end{align}
Here, the Sigmoid function acts elementwise as a global function.
The use of Sigmoid, which is commonly employed in shallow-layer neural networks, makes it well-suited for an alternative as a kernel function.

\vspace{0.2cm}
\noindent(4) \textbf{Tanh kernel} is given by
\begin{align}
    \mathcal{K}_{\mathbf{\Theta}} =
    \mbox{Tanh} \left( \vphantom{G^T_T} \langle\phi(\mathbf{p}; \mathbf{\Theta}), \phi(\mathbf{p}^*; \mathbf{\Theta}) \rangle \right) \,.
\end{align}
Similar to Sigmoid, the Tanh kernel function also globally applies to each element.
However, a notable difference is that the tanh function has a faster rate of changes in values.

\vspace{0.2cm}
\noindent(5) \textbf{Laplacian kernel} function is very similar to the radial basis function, as it also involves applying an exponential function to the distance of two features.
However, the Laplacian kernel is less sensitive to changes in the Gaussian scale parameter $\sigma$.
The expression for the Laplacian kernel is given by

\begin{align}
    \mathcal{K}_{\mathbf{\Theta}} =
    \exp{\left( -\frac{\lVert \phi(\mathbf{p}; \mathbf{\Theta}) - \phi(\mathbf{p}^*; \mathbf{\Theta}) \rVert}{\sigma} \right)} \,. 
\end{align}

Other kernel functions such as polynomial kernel, inverse quadratic kernel,~\etc are not suitable for scene flow tasks and the discussion is beyond the scope of this paper.

\begin{figure}[t]
    \centering
    \includegraphics[width=0.65\linewidth]{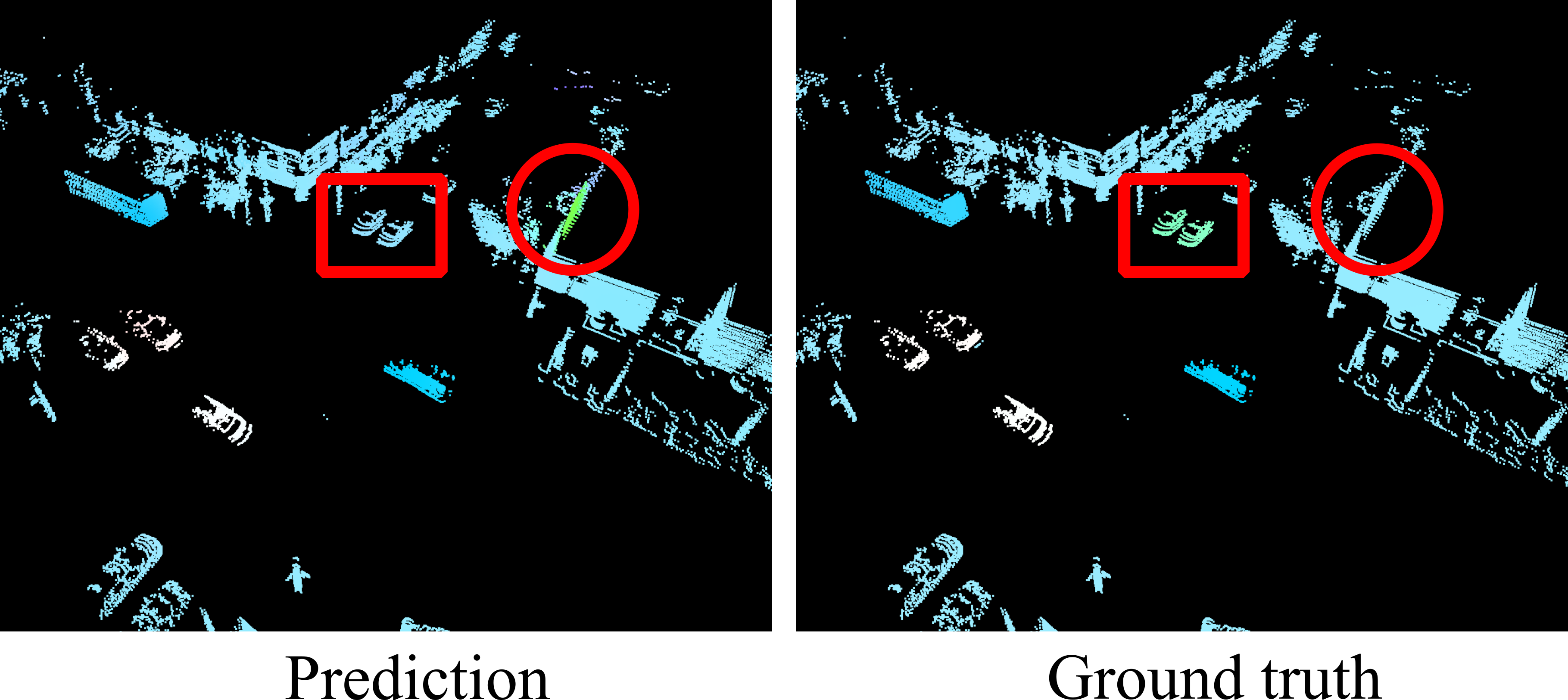}
    \caption{Limitations of our method.
    In this example from Argoverse, while most predicted motions are smooth and accurate, the dynamic cars highlighted in the red box are inaccurately predicted as rigid, similar to the background motion. 
    Additionally, the predicted background motion, marked by the red circle, exhibits extra noise.}
    \label{fig:argo_failure}
    \vspace{-0.3cm}
\end{figure}

\section{Limitations}\label{sec:limitations}
As a linear kernel model, our method has several limitations inherited from the kernel method.
One of the most distinct disadvantages is that when applying learned transformer-based features, it cannot scale up properly because of the big self-attention matrix of the dense point features.
Although we aggregate features from k-nearest neighbors, the computation is still significant.
Further improvement in efficiently using transformer-based point features is needed.

Another limitation lies in the DT loss.
The approximation of the distance transform loss using the grid points can be coarse when applied to scenes that require high fidelity.
It turns out that nearby objects that have different motions may not be easily separated.
A typical failure case is shown in~\cref{fig:argo_failure}.

\section{Conclusion}\label{sec:conclusion}
We present a novel kernel approach to represent large-scale lidar scene flow.
Our kernel method stands out for its simplicity in only requiring solving single-layer linear coefficients during runtime, maintaining great generalizability to out-of-distribution data.
With an implicit embedding of point features, we explore raw points, learned features, and analytical positional encodings as per-point embeddings for a robust kernel representation.
Additionally, our method obtains great efficiency as being a linear model, achieving near real-time performance when tested on dense lidar points.

\appendix

\section{Additional Context}
\label{supp:sec:additional_context}

\subsection{Figure 1 Breakdown}
\begin{figure}
    \centering
    \includegraphics[width=0.6\linewidth]{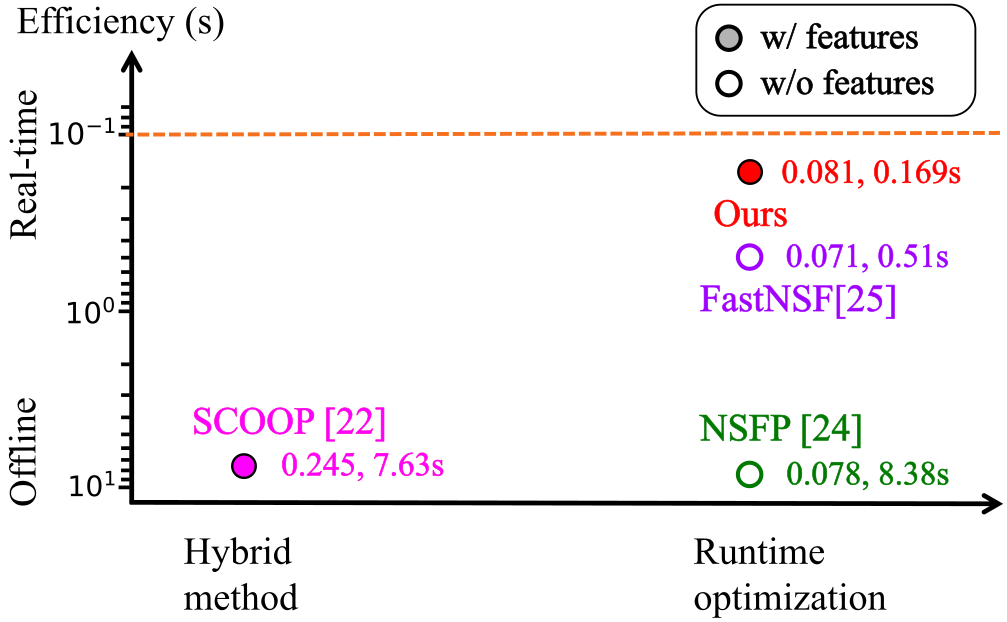}
    \caption{Performance and efficiency comparison of runtime optimization methods and hybrid methods.
    The orange dashed line shows the real-time computation at around 0.1s.
    SCOOP~\cite{lang2023scoop} and our method stand out by utilizing point features---whether learned or analytical PPE features---setting themselves apart from NSFP~\cite{li2021neural} and FastNSF~\cite{li2023fast}, which rely solely on raw coordinate locations without additional features.
    SCOOP is extremely slow and shows inferior performance compared to runtime optimization methods.
    In contrast, our method shows comparable performance (end-point error) with NSFP and FastNSF while being 50 times faster than NSFP.
    }
    \label{supp:fig:figure_01}
\end{figure}

In the main paper, we provide an overview of current scene flow methods with three key properties in~\cref{fig:figure_01}.
To offer a clearer illustration, we further break down the figure, showing the performance of runtime optimization and hybrid methods in~\cref{supp:fig:figure_01}.
Shaded shapes represent methods that utilize point features, whereas hollow legends denote methods that do not leverage any point features.
For instance, SCOOP~\cite{lang2023scoop} relies on learned point features obtained from the correspondence network. 
In contrast, our method with the PEAT-KNN kernel utilizes learned PPE features from point transformer networks, and our method with the RFF-based PE employs analytical PPE features from positional encodings.
We can clearly see that under real-world scenarios with the full point cloud, our method achieves comparable performance while being $\sim$50 times faster than NSFP~\cite{li2021neural}, achieving near real-time performance as compared to feed-forward learning methods.

\subsection{Supervised Learning for PEAT Features}

In~\cref{sec:method:peat} of the main paper, we have proposed to use PEAT~\cite{zheng2023robust}features as learned features for kernel representation.
We propose to use supervised learning for PEAT features when the ground truth flow is provided, for example, in the synthetic dataset FlyingThings3D~\cite{mayer2016large}.

Recall that the scene flow is represented using a per-point embedding-based feature as
\begin{align}
    \mathbf{f} & = \sum_m^M \bm{\alpha}_m \mathcal{K} \left( \phi\left( \mathbf{p}; \mathbf{\Theta} \right), \phi \left( \mathbf{p}^*; \mathbf{\Theta} \right) \right)
    \label{supp:eq:feature_kernel} \\
    & = \sum_m^M \bm{\alpha}_m \mathcal{K}_{\mathbf{\Theta}} \,,
    \label{supp:eq:feature_kernel_simple}
\end{align}
Therefore, the end-to-end supervised learning becomes optimizing both network parameters $\mathbf{\Theta}$ and the linear coefficients $\bm{\alpha}$ as
\begin{align}
    \argmin_{\bm{\alpha}, \mathbf{\Theta}} \left\Vert \mathbf{f}_{gt} - \sum_m^M \bm{\alpha}_m \mathcal{K}_{\mathbf{\Theta}} \right\Vert^2_2 + \lambda \left| \bm{\alpha} \right| \,.
    \label{supp:eq:supervised_learning_01}
\end{align}
Since the optimization of these two parameters is interdependent, one parameter relies on the other to solve, we propose to implicitly embed $\bm{\alpha}$ to relax the optimization.

Suppose we already solved $\mathcal{K}_{\mathbf{\Theta}}$, with the ground truth flow $\mathbf{f}_{gt}$ at hand, we could solve $\bm{\alpha}$ in closed-form using~\cref{supp:eq:feature_kernel_simple} as
\begin{align}
    \bm{\alpha} = \mathbf{f}_{gt} \: \mathcal{K}^{\dagger}_{\mathbf{\Theta}} \,,
    \label{eq:solve_alpha}
\end{align}
and $\dagger$ denotes the Moore–Penrose inverse.
When plugging back into~\cref{supp:eq:supervised_learning_01}, we have the learning objective written as
\begin{align}
    \argmin_{\mathbf{\Theta}} \left\Vert \mathbf{f}_{gt} - \mathbf{f}_{gt} \: \mathcal{K}^{\dagger}_{\mathbf{\Theta}} \: \mathcal{K}_{\mathbf{\Theta}} \right\Vert^2_2 + \lambda \left| \mathbf{f}_{gt} \: \mathcal{K}^{\dagger}_{\mathbf{\Theta}} \right| \,.
    \label{eq:supervised_learning_02}
\end{align}
where $\bm{\alpha}$ is implicitly embedded in.
Since the kernel function might be ill-conditioned or close to singular, we add a small identity matrix to regularize the pseudo-inverse as
\begin{align}
    \argmin_{\mathbf{\Theta}} \left\Vert \mathbf{f}_{gt} - \mathbf{f}_{gt} \: \left( \mathcal{K}_{\mathbf{\Theta}} + \delta \mathbb{I} \right)^{\dagger} \: \mathcal{K}_{\mathbf{\Theta}} \right\Vert^2_2 + \lambda \left| \mathbf{f}_{gt} \: \left( \mathcal{K}_{\mathbf{\Theta}} + \delta \mathbb{I} \right)^{\dagger} \right| \,,
    \label{eq:supervised_learning_03}
\end{align}
where $\mathbb{I}$ is the identity matrix, $\delta$ is a small regularization coefficient.

From~\cref{eq:supervised_learning_03}, we observe that the kernel function is prone to being an identity matrix when minimizing the objective.
To make the optimization more robust, we propose to split the point set into two parts during training.
Specifically, we split the source point cloud $\mathcal{S}_1$ into two sets $\mathcal{A}$ and $\mathcal{B}$.
Accordingly, the ground truth flow is split into set $\mathcal{G}$ and $\mathcal{H}$.
Here is the modified objective:
\begin{align}
    \argmin_{\mathbf{\Theta}} \left\Vert \mathbf{f}_{gt}^{\mathcal{H}
    } - \mathbf{f}_{gt}^{\mathcal{G}} \: \left( \mathcal{K}^{\mathcal{A}}_{\mathbf{\Theta}} + \delta \mathbb{I} \right)^{\dagger} \: \mathcal{K}_{\mathbf{\Theta}}^{\mathcal{B}} \right\Vert^2_2 + \lambda \left| \mathbf{f}_{gt} \left( \mathcal{K}_{\mathbf{\Theta}} + \delta \mathbb{I} \right)^{\dagger} \right| \,.
    \label{eq:supervised_split_data}    
\end{align}
The above objective holds given the fact that $\bm{\alpha}$ only depends on the supporting points while we only split the source point cloud.

\begin{table}[t]
\caption[]{\textbf{Performance on Waymo Open scene flow dataset.}
We tested with 202 scene flow examples from the Waymo Open validation dataset, presenting the results in two sections. 
The upper table shows results for the full point cloud, with the number of points ranging approximately from 8k to 144k.
In this section, we focus on discussing runtime optimization-based methods and hybrid methods.
The lower table shows results for a subset of the dataset, limited to 8,192 points in each sample, where we specifically include results for learning-based methods.
In both tables, performance levels are color-coded as: \colorbox{nord_green}{green box} denotes the best level of performance, \colorbox{nord_yellow}{yellow box} represents the second-best level of performance, and \colorbox{nord_red}{red box} indicates the worst level of performance.
We categorize each method as either a runtime optimization \cmark, a hybrid \faCircleO, or a feed-forward learning method \xmark.
Additionally, we indicate whether the method leverages point features as \cmark or \xmark.
}
    \centering
    \begin{adjustbox}{width=\linewidth}
    \begin{tabular}{@{}clccx{0.13\linewidth}x{0.13\linewidth}x{0.13\linewidth}x{0.13\linewidth}x{0.13\linewidth}c@{}}
        \toprule
        & \thead{\normalsize Method} & \thead{\normalsize Runtime Opt.} & \thead{\normalsize Point Features} &\thead{${\mathcal{E}}{(m)}\downarrow$} &\thead{${Acc_5}{(\%)}\uparrow$} &\thead{${Acc_{10}}{(\%)}\uparrow$} &\thead{${\theta_{\epsilon}}{(rad)}\downarrow$} &\thead{$T(s)\downarrow$} &\\
        \midrule
        & NSFP~\cite{li2021neural} (baseline) & \cmark & \xmark & 0.118 & \cellcolor{nord_yellow}74.16 & \cellcolor{nord_yellow}86.70 & \cellcolor{nord_yellow}\underline{0.300} & \cellcolor{nord_red}18.39 \\
        & SCOOP~\cite{lang2023scoop} & \faCircleO & \cmark & \cellcolor{nord_red}0.201 & \cellcolor{nord_red}61.07 & \cellcolor{nord_red}76.07 & \cellcolor{nord_red}0.384 & \cellcolor{nord_red}21.68 \\
        & FastNSF~\cite{li2023fast} & \cmark & \xmark & \cellcolor{nord_green}\textbf{0.072} & \cellcolor{nord_green}\textbf{84.73} & \cellcolor{nord_green}\textbf{92.24} & \cellcolor{nord_green}\textbf{0.280} & \cellcolor{nord_yellow}0.58 \\
        & FastNSF~\cite{li2023fast} (linear) & \cmark & \xmark & \cellcolor{nord_yellow}0.109 & 71.27 & \cellcolor{nord_yellow}85.80 & 0.321 & \cellcolor{nord_yellow}0.49 \\
        & Ours (Point, CD) & \cmark & \xmark & 0.115 & 70.05 & \cellcolor{nord_yellow}84.26 & 0.313 & 3.24 \\
        & Ours (Point, DT) & \cmark & \xmark & 0.115 & \cellcolor{nord_yellow}73.93 & \cellcolor{nord_yellow}\underline{86.72} & 0.317 & \cellcolor{nord_green}\textbf{0.153} \\
        & Ours (PEAT-KNN, CD) & \faCircleO & \cmark & 0.131 & \cellcolor{nord_red}62.22 & 81.15 & 0.317 & \cellcolor{nord_red}22.97 \\
        & Ours (RFF, CD) & \cmark & \cmark & \cellcolor{nord_yellow}\underline{0.100} & 71.60 & \cellcolor{nord_yellow}85.50 & \cellcolor{nord_yellow}0.302 & 4.39 \\
        & Ours (RFF, DT) & \cmark & \cmark & \cellcolor{nord_yellow}0.110 & \cellcolor{nord_yellow}\underline{75.31} & \cellcolor{nord_yellow}86.34 & \cellcolor{nord_yellow}0.303 & \cellcolor{nord_green}\underline{0.221} \\
        \midrule
        & FLOT~\cite{puy20flot} & \xmark & \cmark & \cellcolor{nord_red}0.702 & \cellcolor{nord_red}2.46 & \cellcolor{nord_red}11.30 & \cellcolor{nord_red}0.808 & \cellcolor{nord_green}0.099 \\
        & R3DSF~\cite{gojcic2021weakly} & \xmark & \cmark & \cellcolor{nord_red}0.414 & \cellcolor{nord_red}35.47 & \cellcolor{nord_red}44.96 & \cellcolor{nord_red}0.527 & \cellcolor{nord_green}\underline{0.140} \\
        & NSFP~\cite{li2021neural} & \cmark & \xmark & 0.138 & 53.62 & \cellcolor{nord_yellow}78.57 & \cellcolor{nord_yellow}\underline{0.339} & 2.46 \\
        & SCOOP~\cite{lang2023scoop} & \faCircleO & \cmark & 0.303 & 41.13 & 64.69 & 0.471 & \cellcolor{nord_yellow}0.278 \\
        & FastNSF~\cite{li2023fast} & \cmark & \xmark & \cellcolor{nord_green}\textbf{0.106} & \cellcolor{nord_green}\textbf{77.53} & \cellcolor{nord_green}\textbf{88.99} & \cellcolor{nord_green}\textbf{0.329} & \cellcolor{nord_green}\textbf{0.121} \\
        & Ours (Point, CD) & \cmark & \xmark & 0.163 & 45.84 & 70.01 & 0.381 & 0.836 \\
        & Ours (Point, DT) & \cmark & \xmark & 0.173 & 50.47 & 72.79 & 0.388 & \cellcolor{nord_green}0.163 \\
        & Ours (PEAT, CD) & \faCircleO & \cmark & 0.168 & 48.67 & 74.00 & 0.371 & 1.52 \\
        & Ours (PEAT, DT) & \faCircleO & \cmark & 0.151 & 49.18 & 74.73 & \cellcolor{nord_yellow}0.354 & \cellcolor{nord_yellow}0.254 \\
        & Ours (RFF, CD) & \cmark & \cmark & \cellcolor{nord_yellow}0.124 & \cellcolor{nord_yellow}56.56 & \cellcolor{nord_yellow}77.64 & \cellcolor{nord_yellow}0.347 & 0.531 \\
        & Ours (RFF, DT) & \cmark & \cmark & \cellcolor{nord_yellow}\underline{0.123} & \cellcolor{nord_yellow}\underline{59.99} & \cellcolor{nord_yellow}\underline{80.39} & \cellcolor{nord_green}\textbf{0.329} & \cellcolor{nord_green}0.152 \\
        \bottomrule
    \end{tabular}
    \label{supp:tab:main_waymo}
    \end{adjustbox}
\end{table}

\section{Additional Results}
\label{supp:sec:additional_experiments}

\begin{figure}[t]
\centering
    \includegraphics[width=\linewidth]{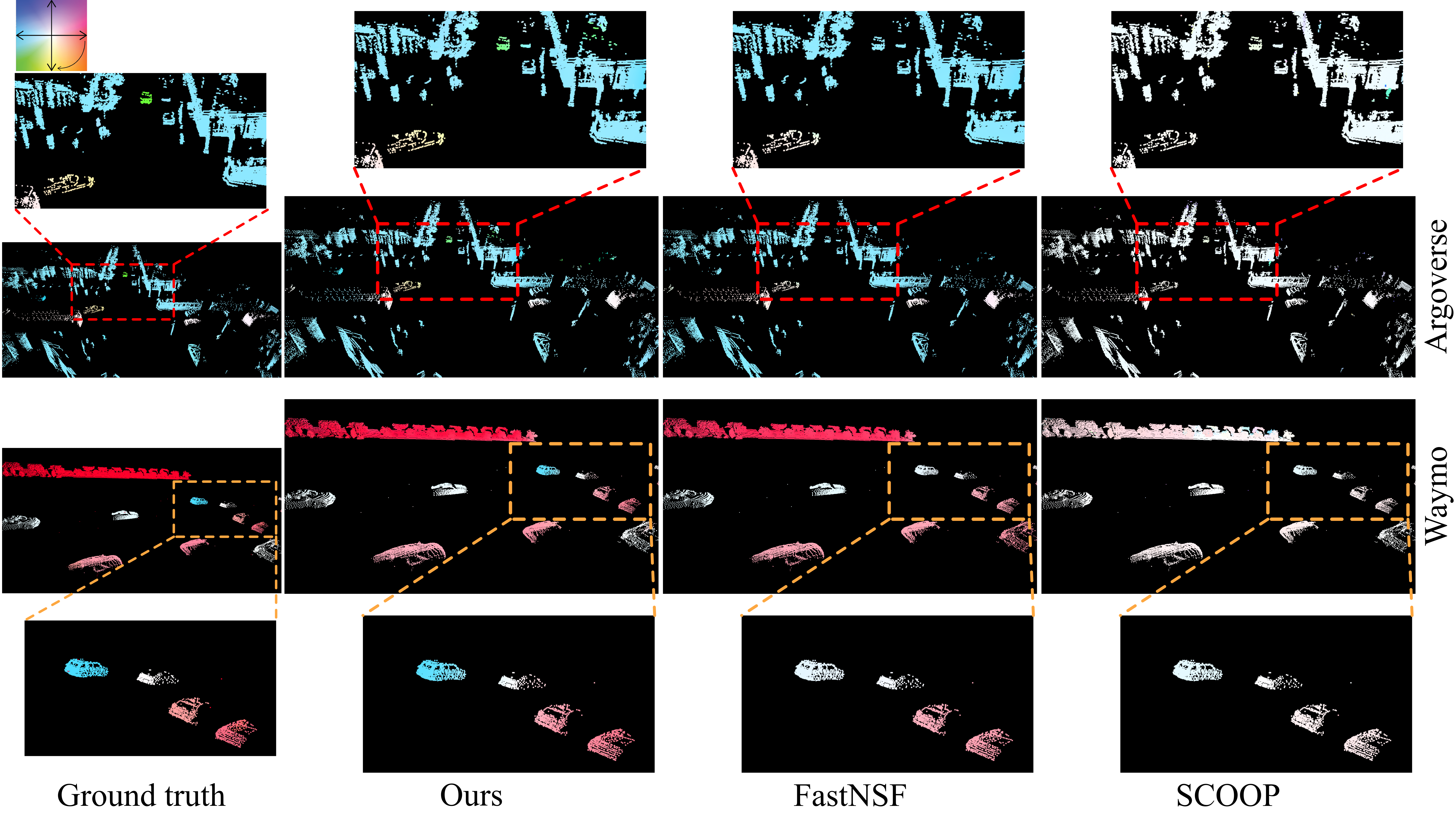}
    \caption{Visual results demonstrate the effectiveness of our method in comparison to FastNSF~\cite{li2023fast} and SCOOP~\cite{lang2023scoop} on two samples from the Argoverse and Waymo Open scene flow dataset. 
    The 3D scene flow is presented in a projected 2D view for clearer illustration.
    In the upper left corner, a color wheel is used to indicate the projected flow magnitude (color intensity) and flow direction (angle). 
    The zoomed-in details of boxed areas are shown here.
    Our method is robust to dynamic AV scenes, successfully capturing both the rigid pose and the dynamic motions.
    FastNSF struggles to capture multiple dynamic objects in some cases, such as the entire scene in the upper row, and the blue car in the bottom row, leading to occasional noisy results.
    SCOOP, on the other hand, cannot scale up well to dense points, resulting in near-rigid estimation for most of the scenes.
    } 
    \label{supp:fig:waymo_qualitative}
\end{figure}

\subsection{Results on Waymo Open Dataset}
\label{supp:sec:waymo_results}
We present additional results on the Waymo Open scene flow dataset~\cite{sun2020scalability}.
The main result is shown in~\cref{supp:tab:main_waymo} and~\cref{supp:fig:waymo_qualitative}.

Here, we observe a similar performance trend compared to results on the Argoverse scene flow dataset as discussed in Sec.~{\color{red}4.1}.

We compared our method to runtime optimization-based methods NSFP~\cite{li2021neural}, FastNSF~\cite{li2023fast}, hybrid method SCOOP~\cite{lang2023scoop}, and learning-based methods FLOT~\cite{puy20flot}, R3DSF~\cite{gojcic2021weakly} on performance and computational efficiency.
In terms of error and accuracy on full lidar points, our approach, along with runtime optimization-based methods NSFP and FastNSF, outperforms hybrid methods SCOOP and our PEAT-KNN kernel method.
Despite the high accuracy of FastNSF, when visualizing in~\cref{supp:fig:waymo_qualitative}, some specific scenes show unreasonable performance.
SCOOP struggles with poor generalizability. 
In contrast, our method achieves competitive performance to NSFP and FastNSF. 
Analytical RFF PE-based features further improve the performance of our method, emphasizing the advantage of leveraging point features. 
Regarding computation time, NSFP and SCOOP are extremely slow.
While the computation overheads in NSFP arise from Chamfer loss and deep network structure, SCOOP encounters challenges in feature extraction and flow refinement.
FastNSF addresses the inefficiency of NSFP with a distance transform loss. 
Remarkably, our method achieves near real-time performance and a substantial speedup compared to NSFP ($\sim$ 90 times faster). 
When compared to learning-based methods FLOT and R3DSF on 8,192 points, our runtime optimization-based approach maintains lower error and higher accuracy, highlighting its advantage in generalizability. 
Adding point features poses challenges for end-to-end trained models, but our hybrid approach yields reasonable and improved accuracy. 
Overall, our method demonstrates comparable efficiency, emphasizing its potential for real-time applications in autonomous driving scenarios.

\begin{figure}[t]
    \centering
    \includegraphics[width=0.8\linewidth]{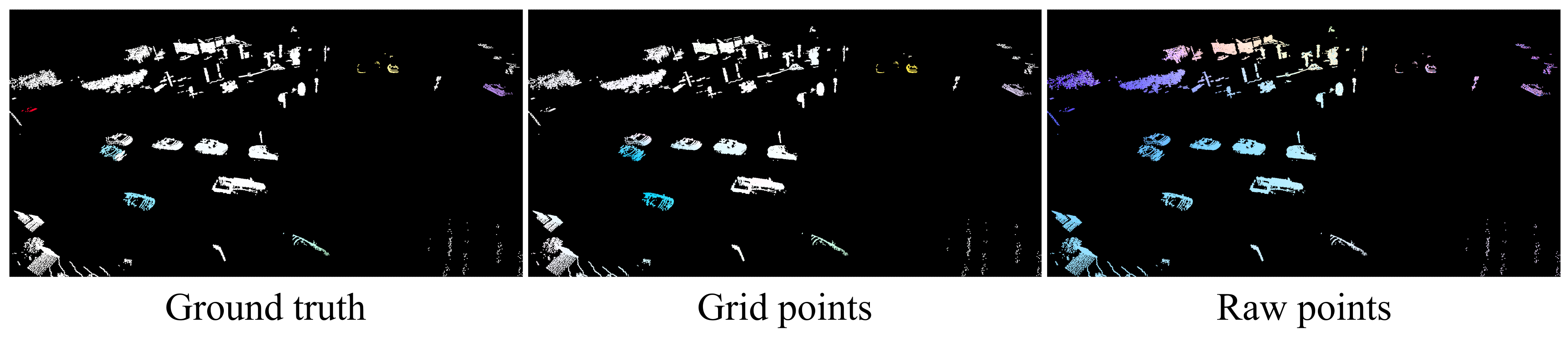}
    \caption{Visual comparison on Waymo Open dataset using RFF-based kernel methods on grid points versus raw points.
    It is important to note that both methods use the full point cloud with dense points.
    Using grid points is robust to noisy data while using raw points results in noisy estimated motions in the background, as well as for nearby objects.
    For example, there are 7 cars in the middle of the scene, where 2 cars are moving toward the west (as compared to the AV), and the other 5 cars are moving at a similar speed/direction as the AV.
    However, when utilizing the raw points, all 7 cars are estimated as moving toward the west, yielding noisy results.
    In contrast, when employing the grid point strategy, the 5 nearby cars are hardly influenced by those 2 cars, showcasing the advantage of using grid points in handling dense point clouds.
    }
    \label{supp:fig:compare_grid_raw_points}
\end{figure}

\subsection{Grid Points vs. Raw Points}
So far, our experiments have primarily employed the grid point strategy proposed in Sec.~{\color{red}3.2}, with the exception of experiments involving PEAT/PEAT-KNN kernels, where features were trained on raw points. 
The preference for grid points is motivated by the dense and noisy nature of raw points, while grid points offer sparsity and separability, demonstrating robustness against noise and having advantages in terms of memory and computational efficiency.

On the other hand, when the point cloud is sparse, the strategy of using grid points does not stand out, as using raw points still maintains relatively efficient computation.

We show a visual comparison of using grid points versus raw points in~\cref{supp:fig:compare_grid_raw_points}, which demonstrates the advantages of using grid points in dense point clouds.

%
%
\bibliographystyle{splncs04}
\bibliography{main}

\end{document}